\documentclass{article}

    \usepackage[preprint, nonatbib]{neurips_2021}

\usepackage[utf8]{inputenc} % allow utf-8 input
\usepackage[T1]{fontenc}    % use 8-bit T1 fonts
\usepackage{hyperref}       % hyperlinks
\usepackage{url}            % simple URL typesetting
\usepackage{booktabs}       % professional-quality tables
\usepackage{amsfonts}       % blackboard math symbols
\usepackage{nicefrac}       % compact symbols for 1/2, etc.
\usepackage{microtype}      % microtypography
\usepackage{xcolor}         % colors

\usepackage{graphicx}
\usepackage{subfig}
\usepackage{amsmath}
\usepackage{amsthm}
\usepackage{booktabs}
\usepackage{algorithm}
\usepackage{algorithmic}

\title{Distribution Preserving Graph Representation Learning}

\author{%
  Chengsheng Mao, Yuan Luo \\
  Department of Preventive Medicine\\
  Feinberg School of Medicine, Northwestern University\\
  Chicago, IL, 60611 \\
  \texttt{\{chengsheng.mao, yuan.luo\}@northwestern.edu} \\
}

\begin{document}

\maketitle

\begin{abstract}
  Graph neural network (GNN) is effective to model graphs for distributed representations of nodes and an entire graph. Recently, research on the expressive power of GNN attracted growing attention. A highly-expressive GNN has the ability to generate discriminative graph representations. However, in the end-to-end training process for a certain graph learning task, a highly-expressive GNN risks generating graph representations overfitting the training data for the target task, while losing information important for the model generalization. In this paper, we propose Distribution Preserving GNN (DP-GNN) - a GNN framework that can improve the generalizability of expressive GNN models by preserving several kinds of distribution information in graph representations and node representations. Besides the generalizability, by applying an expressive GNN backbone, DP-GNN can also have high expressive power. We evaluate the proposed DP-GNN framework on multiple benchmark datasets for graph classification tasks. The experimental results demonstrate that our model achieves state-of-the-art performances. 
\end{abstract}

\section{Introduction}
Graph structured-data plays an important role in describing the relationships between objects and has been widely utilized in modeling diverse real-world datasets including chemical compounds, molecular graph structures, protein-protein interaction networks, and social networks. Representation learning with graph structured-data usually requires one to learn effective representations that capture the graph structures as well as the features of nodes and edges. Recently, Graph Neural Network (GNN) attracted growing attention in graph representation learning in various domains such as text mining \cite{yao2019graph}, clinical decision making \cite{li2018classifying,mao2022medgcn}, and image processing \cite{mao2022imagegcn,garcia2017few}. GNN updates the node representation recursively by aggregating the neighborhood information of the node  \cite{hamilton2017inductive,morris2019weisfeiler,xu2019powerful,mao2020towards}. Finally, the representation of an entire graph is obtained through aggregating the node representations in the graph. The final node or graph representations can be fed into downstream learning tasks for end-to-end training. 

Recently, researchers paid growing attention to improving the expressive power of GNNs, i.e., to obtain the representations that can effectively discriminate non-equivalent nodes or non-isomorphic graphs. In GNN, the aggregation rule plays a vital role in learning expressive representations for the nodes and the entire graph \cite{xu2019powerful,mao2020towards}. While various aggregation rules in GNN were proposed to achieve good performances in different tasks \cite{kipf2017semi,hamilton2017inductive,zhang2018end,xinyi2019capsule,wang2020haar}, \cite{xu2019powerful} indicated that GNNs can at most achieve the expressive power as the Weisfeiler-Lehman graph isomorphism test (WL test) and developed Graph Isomorphism Network (GIN) that can achieve the expressive power of the WL test by designing an injective neighborhood aggregation function and a graph-level readout function. 

However, if a GNN possess a high expressive power, it would have a high risk of overfitting the training data when trained for a certain graph learning task, because it is likely to generate quite different distributions for graph or node representations between training data and test data. In this paper, we propose a GNN framework, Distribution Preserving GNN (DP-GNN), to alleviate the overfitting issue by preserving several kinds of distribution information in the node and graph representations. In the proposed framework, besides the major graph classification task, we also adopt 4 auxiliary tasks for distribution preserving. By the regularization of multiple distribution preserving tasks, DP-GNN is expected to achieve a better generalizability.

\begin{figure}
    \centering
    \includegraphics[width=0.8\linewidth,page=2]{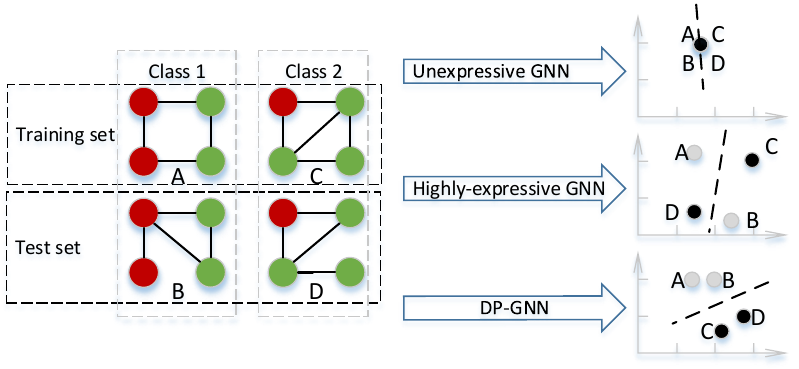}
    \caption{An example illustrating the differences between different GNNs in graph representations. With different colors indicating different node labels, the 4 graphs for classification are non-isomorphic to each other. Unexpressive GNN may map different graphs to the same representation. highly-expressive GNN can map different graphs to different representations, but the model trained on training data (A and C) cannot be generalized to test data (B and D). Our framework DP-GNN can map different graphs to different representations, and can have better generalizability. }
    \label{fig:example}
\end{figure}

Figure \ref{fig:example} shows a toy example to illustrate the differences manifestations between unexpressive GNN, highly-expressive GNN, and DP-GNN in graph representation. In Figure \ref{fig:example}, the 4 graphs for classification are non-isomorphic to each other, A and C in training set, B and D in test set. The unexpressive GNN generates the same representations for different graphs, thus the classification task fails. While highly-expressive GNN can generate quite different representations for different graphs, a line that can separate A and C cannot correctly separate B and D, which implies the model trained on training data (A and C) cannot be generalized to test data (B and D). For the proposed DP-GNN, because the generated graph representation preserves the multiple kinds of distribution information which is beneficial to the graph classification task, it can not only generate different representations for different graphs, but also the graph representations on test data can be correctly separated by a line separating the training data, implying a better generalizability. Our contributions are summarized as follows: 
\begin{itemize}
    \item We propose a distribution preserving GNN framework to improve the generalizability of expressive GNNs. To the best of our knowledge, this is the first study to explore the generalizability of expressive GNNs by distribution preserving.
    \item We adopt several auxiliary distribution preserving tasks to regularize the training process. By tuning the weight of different tasks, we can obtain the importance of different auxiliary tasks for graph classification task for different problems.
    \item The experimental results on several benchmark datasets for graph classification demonstrate that our model achieves state-of-the-art performances. Also, the experimental results validate the generalizability and expressive power of our model.
\end{itemize}

\section{Related work}

\subsubsection{Expressive Power of GNN.}
Most modern GNNs were instances of the family of message passing neural networks \cite{gilmer2017neural} that follow a neighborhood aggregation strategy that recursively updates the representation of a node by aggregating representations of its neighbors and the node itself. The graph-level representation is obtained through aggregating the final representations of all the nodes in the graph. Recently, the expressive power of GNNs attracted increasing attentions. The expressive power of GNNs describes how a GNN model can distinguish different graphs or nodes. Most of the previous GNNs are designed based on empirical intuition, heuristics, or experimental trial-and-error until \cite{xu2019powerful} presented a theoretical framework for analyzing the expressive power of GNNs. \cite{xu2019powerful} showed that message passing GNNs can at most achieve the expressive power of the 1-WL test and developed GIN that can achieve this expressive power. By the conclusion of \cite{xu2019powerful}, to achieve expressive GNNs, it is critical to design an injective neighborhood aggregation function and an injective graph-level aggregation function. Inspired by the study of \cite{xu2019powerful}, \cite{mao2020towards} proposed to improve the expressiveness of GNN for attributed graphs by designing continuous injective set functions for neighborhood aggregation. \cite{maron2019provably} developed a GNN model that incorporates standard multi-layer perceptrons (MLPs) of the feature dimension and a matrix multiplication layer, possessing the expressiveness of 3-WL. A survey on the expressive power on GNN can be found in \cite{sato2020survey}. 

Although an expressive GNN model has the ability to learn distinguishing graph representations, it has a high risk overfitting the training data in the end-to-end training for a certain graph learning task. While more and more studies focused on improving the expressiveness of GNNs, few studies in the literature were found to consider the generalizability of expressive GNNs. In this paper, we try to improve the generalizability of GNN by distribution preserving.

\subsubsection{Multi-task Learning.}
Multi-task learning (MTL) is a machine learning paradigm that jointly optimizes the model for multiple learning tasks. By leveraging the information contained in multiple related tasks, an MTL model is expected to learn each task more accurately.  A machine learning model optimized for a single task on training data may loss some relevant information captured in other related tasks, which could assist in generalizing the predictions to unseen data. Technically, MTL can help improve the generalization of models by introducing an inductive bias \cite{zhang2017survey}. A survey on MTL can be found in \cite{zhang2017survey}. Some works also applied MTL to graph representation learning. \cite{mao2022medgcn} employed a graph convolutional network to jointly learn two medical tasks, medication recommendation and lab test imputation. \cite{holtz4multi,xie2020multi} employed a GNN to jointly learn a node-level prediction task and a graph-level prediction task. \cite{wang2020multi} proposed a MTL framework for network embedding based on two tasks that respectively preserve the global and local structural information. \cite{xie2020multitask} presented a MTL framework that simultaneously performs multiple tasks including node classification and link prediction. However, the task of distribution preserving in graph representation has not been explored in-depth in the literature. This paper will consider the distribution preserving tasks for graph representation learning.

\section{Methods}
\subsection{Preliminaries}
\subsubsection{Problem Statement}
A graph $G$ is denoted as $(V,E)$, where $V$ is the node list (assume size $n$) corresponding to a node feature matrix $X\in {R}^{n\times d}$ and a list of node labels $L=[l_1,\cdots,l_n], l_i \in \{0,\cdots,C_N-1\}$ ($C_N$ is the number of classes of nodes), and $E$ is the set of edges corresponding an adjacency matrix $A\in\{0,1\}^{n\times n}$. For a set of labeled graphs $D = \{(G_1, y_1), \cdots, (G_N, y_N)\}$, where $y_i\in \{0,\cdots,C_G-1\}$ is the graph-level label associated with graph $G_i$, $C_G$ is the number of classes of graphs. For graph classification, we aim to learn a model $f(\cdot)$ that is able to correctly predict the graph label with the input of the graphs in $D$, i.e., $y=f(G),  \mbox{for } (G,y)\in D$. In this paper, $[x,y]$ denotes the concatenation of two vectors $x$ and $y$ or a vector with two elements $x$ and $y$. $\mathbf{x}[i]$ denotes the $i$th entry of vector $\mathbf{x}$.   

\subsubsection{Expressive Graph Neural Networks}
An expressive GNN should be able to learn node or graph representations that incorporate information of the graph structure and node features, which are vital to discriminate different graphs. Most modern GNNs fall into the category of message passing GNNs. To get a node's representation, a message passing GNN first aggregates the node's neighbors' representations to achieve the neighborhood representation, and then combine the neighborhood representation and the node's current representation to achieve the new representation of the node. By 1 iteration, a node representation should contains information from its 1-hop neighborhood. After \textit{k} iterations of aggregation, a node representation captures the structural information of its \textit{k}-hop neighborhood. In this paper, unless otherwise stated, GNN defaults to a message passing GNN. Formally, the propagation rule of a GNN layer can be represented as
\begin{equation}\label{eq:aggregate}
H_{\mathcal{N}(v)}^{(k)} = f_{A}^{(k)}\left( \left\{H^{(k)}(w) | w\in \mathcal{N}(v) \right\}\right)
\end{equation}
\begin{equation}\label{eq:combine}
H^{(k+1)}(v) =    
f_{C}^{(k)} \left(H^{(k)}(v) ,H_{\mathcal{N}(v)}^{(k)} \right)  
\end{equation}
where $H^{(k)}(v)$ is the representation vector of node $v$ in the $k$th layer, and $H^{(0)}(v)$ is initialized with $X(v)$, the original feature vector of node $v$. $\mathcal{N}(v)$ is the neighborhood of $v$. $f_{A}^{(k)}(\cdot)$ is an aggregation function over a set of node representations; Eq. \ref{eq:aggregate} aggregates the neighbors' representations to achieve the neighborhood representation. Eq. \ref{eq:combine} combines the node's current representation and its neighborhood’s representation in the $k$th layer to achieve the new node representation. Another aggregation function $f_{R}(\cdot)$ is employed to obtain the graph-level representation $H_G$ by aggregating the final representations of all nodes in the graph $G$, for a $K$-layer GNN, i.e.,
\begin{equation}\label{eq:readout}
H_G = f_{R} \left( \left\{H^{(K)}(v) | v\in G \right\}) \right)
\end{equation}

$H_G$ can be fed to a classifier for graph classification. To achieve a high expressive power for a GNN, $f_{A}(\cdot)$, $f_{C}(\cdot)$ and $f_{R}(\cdot)$ all should be injective \cite{xu2019powerful}. An injective function for $f_C(\cdot)$ that operates on two vectors can be easily achieved by concatenating the two vectors. Following the conclusion in \cite{mao2020towards}, an injective set function for $f_A(\cdot)$ or $f_R(\cdot)$ can be achieved by summing a certain transformation function over each element over the set, i.e.,
\begin{equation}\label{eq:poweragg}
f_{A}^{(k)}\left( \left\{H^{(k)}(w) | w\in \mathcal{N}(v) \right\}\right) = \sum_{w\in \mathcal{N}(v)}{\Phi^{(k)} \left(H^{(k)}(w) \right)}
\end{equation}
\begin{equation}\label{eq:readoutphi}
f_{R} \left( \left\{H^{(K)}(v) | v\in G \right\} \right) = \sum_{v\in G}{\Phi_G \left(H^{(K)}(v)\right)}
\end{equation}
where $\Phi^{(k)}(\cdot)$ and $\Phi_G$ are certain transformation functions that make the set function $f_A(\cdot)$ and $f_R(\cdot)$ continuous and injective, respectively.

\subsection{DP-GNN}
We adopt the expressive GNN architecture with learnable transformations in \cite{mao2020towards} as our backbone, where  $\Phi^{(k)}(\cdot)$ and $\Phi_G$ are approximated by learnable MLPs. Our propagation rule is
\begin{equation}\label{eq:MLPagg}
H^{(k+1)}(v)=   
\mbox{MLP}_{C}^{(k)} \left( \left[H^{(k)}(v), \sum_{w\in \mathcal{N}(v)}{\mbox{MLP}_{T}^{(k)}\left(H^{(k)}(w)\right)}\right] \right)
\end{equation}
\begin{equation}\label{eq:readoutmlp}
H_G = \sum_{v\in G}{\mbox{MLP}_G \left(H^{(K)}(v)\right)}
\end{equation}
where $\mbox{MLP}_{T}^{(k)}$ and $\mbox{MLP}_{C}^{(k)}$ are the transformation function for neighborhood aggregation and the combine function for the \textit{k}th layer, respectively. $\mbox{MLP}_G$ is the transformation function for graph-level aggregation.

\begin{figure*}
    \centering
    \includegraphics[width=\linewidth,page=1]{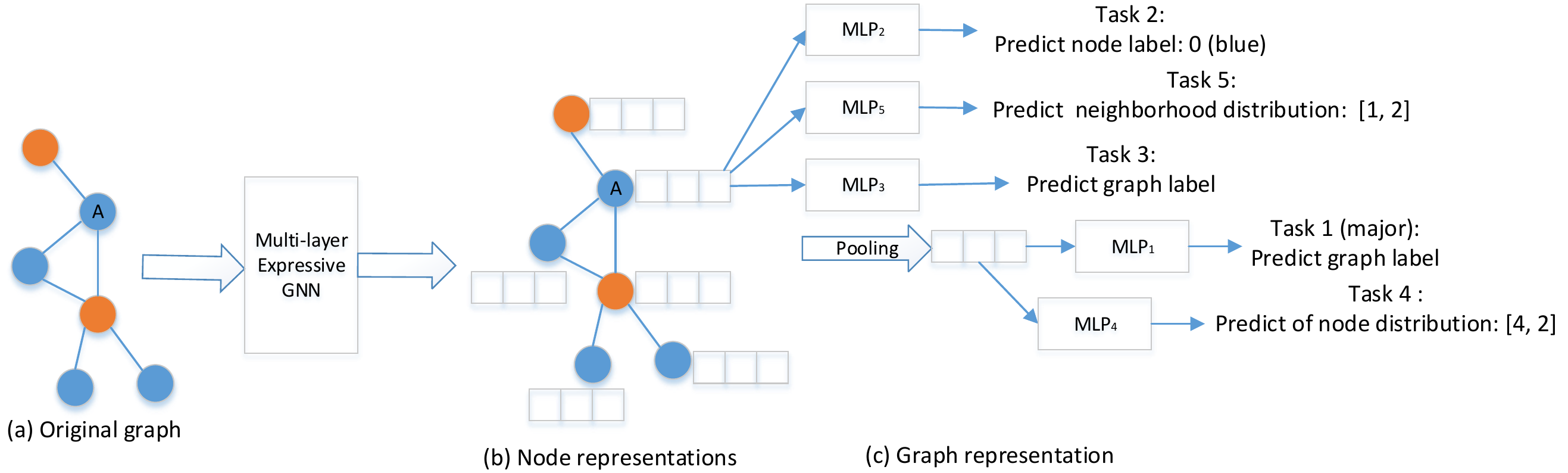}
    \caption{The overview of DP-GNN. An original graph gets its node representations through a multi-layer expressive GNN, and the node representations are aggregated to get the graph representation. Multiple tasks are used to constrain the node or graph representations. Task 1 is the major graph classification task; task 2 is to preserve the node label into node representation; task 3 is to preserve the graph label into node representations; task 4 is to preserve the information of node type distribution of the entire graph into the graph representation (i.e., [4,2] for 4 blue nodes and 2 yellow nodes in the graph); task 5 is to preserve the information of neighborhood distribution of a node into the node representation (i.e., [1,2] for one blue nodes and 2 yellow nodesin the neighborhood of A).}
    \label{fig:framework}
\end{figure*}

By the above propagation rule, an expressive GNN almost maps different graphs to different representations. Due to the different representations for different graphs, the representations of training graphs and test graphs are of high risk encountering distribution shift. A model optimized for a single task on the training set may overfit the training set. Since MTL can improve the generalization performance of multiple tasks when they are related \cite{zhang2017survey}, we use a MTL framework to achieve more generalized graph representations. Besides the major graph classification task, we set 4 auxiliary tasks for distribution preserving as a regularization method. The overview of the proposed DP-GNN model is illustrated in Figure \ref{fig:framework}, where an original graph gets its node representations through a multi-layer expressive GNN with propagation rule Eq. \ref{eq:MLPagg}, and the node representations are aggregated to get the graph representation by Eq. \ref{eq:readoutmlp}. We detail the tasks in Section \ref{sec:tasks}.

\subsection{Tasks for Distribution Preserving}\label{sec:tasks}
In this section, we introduce the major task and the 4 auxiliary tasks each of which corresponds to some distribution information in the graph.

\subsubsection{Major task: graph classification}
For a graph $G$, the final graph representation $H_G(G)$ is fed to an MLP for graph classification. For a total of $C_G$ classes of graphs, the MLP outputs a vector $S(G)$ of length $C_G$ corresponding to the support of the $C_G$ classes, i.e., $S(G) = \mbox{MLP}_1(H_G(G))$. In test process, the index of maximum $C_G$ is considered as the graph label. In training process, for the true label $y(G)$, we get the cross-entropy loss as 
\begin{equation}
    \mathcal{L}_1(G) = -S(G)[y(G)]+\log \left(\sum_{i=0}^{C_G-1}\exp{(S(G)[i])} \right)
\end{equation}
\subsubsection{Task 2: node label preserving}
Usually, the graph label is related to the node label, thus a node representation that preserves the node label can be aggregated to an informative graph representation that helps the graph classification. Node label preserving is actually to preserve the probability distribution of a node's label in the node representation. For a node $v$, the final node representation $H(v)$ is fed to an MLP for node label prediction. For a total of $C_N$ classes of nodes, the MLP outputs a vector $S_N(v)$ of length $C_N$ corresponding to the support of the $C_N$ classes, i.e., $S_N(v) = \mbox{MLP}_2(H(v))$. In training process, for the true label $l(v)$, we get the cross-entropy loss as 
\begin{equation}
    \mathcal{L}_2(v) = -S_N(v)[l(v)]+\log \left(\sum_{i=0}^{C_N-1}\exp{(S_N(v)[i])} \right)
\end{equation}

\subsubsection{Task 3: graph label preserving}
In a $k$-layer GNN, a node representation captures the information of its $k$-hop neighborhood, thus, for a graph with longest path not longer than $k$, each node representation can capture the information of the entire graph. Preserving the graph label in the node representation could be helpful to the graph classification. Graph label preserving is actually to preserve the probability distribution of a graph's label in the node representation. For a node $v$ in graph $G$, the final node representation $H(v)$ is fed to an MLP for graph label prediction. For a total of $C_G$ classes of graphs, the MLP outputs a vector $S_G(v)$ of length $C_G$ corresponding to the support of the $C_G$ classes, i.e., $S_G(v) = \mbox{MLP}_3(H(v))$. In training process, for the true label $y(G)$ of graph $G$, the cross-entropy loss is 
\begin{equation}
    \mathcal{L}_3(v) = -S_G(v)[y(G)]+\log \left(\sum_{i=0}^{C_G-1}\exp{(S_G(v)[i])} \right)
\end{equation}

\subsubsection{Task 4: graph node distribution preserving}
Graph node distribution refers to the number of nodes of each type contained in the graph. The node distribution in a graph could affect the class of a graph. For example, the two classes in Figure \ref{fig:example} have different node distributions in a graph, Class 1 contains 2 red and 2 green nodes, while Class 2 contains 1 red and 3 green nodes. Preserving graph node distribution in the graph representation could help graph classification. 
For a graph $G$, the final graph representation $H_G(G)$ is fed to an MLP for node distribution prediction. For a total of $C_N$ classes of nodes, the MLP outputs a vector $P_G(G)$ of length $C_N$ corresponding to the number of nodes in each of the $C_N$ classes, i.e., $P_G(G) = \mbox{MLP}_4(H_G(G))$. In training process, we can count the number of nodes of each type in the graph $G$ as the true node distribution $D_G(G)$, we get the mean squared error (MSE) loss as 
\begin{equation}
    \mathcal{L}_4(G) = \frac{1}{C_N}\sum_{i=0}^{C_N-1} \left(P_G(G)[i]-D_G(G)[i] \right)^2
\end{equation}

\subsubsection{Task 5: neighborhood distribution preserving}
For a node $v$ in graph $G$, the neighborhood distribution refers to how many nodes of each type can be found in the neighbors of $v$ in graph $G$. Node representations that preserve the neighborhood distribution would help the aggregated graph representation aware of the neighborhood of each node. Thus, if the graph classification is related to the nodes' neighborhood, this task is helpful. For a node $v$ in graph $G$, the final node representation $H(v)$ is fed to an MLP for neighborhood distribution prediction. For a total of $C_N$ classes of nodes, the MLP outputs a vector $P_N(v)$ of length $C_N$ corresponding to the number of nodes in each of the $C_N$ classes, i.e., $P_N(v) = \mbox{MLP}_5(H(v))$. In training process, we can count the number of nodes of each type in the neighborhood of $v$ as the true neighborhood distribution $D_N(v)$, we get the MSE loss as 
\begin{equation}
    \mathcal{L}_5(v) = \frac{1}{C_N}\sum_{i=0}^{C_N-1} \left(P_N(v)[i]-D_N(v)[i] \right)^2
\end{equation}

\subsection{Training Strategy}
We train the model by mini-batch, for a mini-batch of graphs $\mathcal{G}$, the graph-level losses ($\mathcal{L}_1$ and $\mathcal{L}_4$) are averaged over all the graphs in $\mathcal{G}$ and the node-level losses ($\mathcal{L}_2$, $\mathcal{L}_3$ and $\mathcal{L}_5$) are averaged over all the nodes in $\mathcal{G}$, i.e., 
\begin{equation}
    \mathcal{L}_1(G) = \frac{1}{|\mathcal{G}|}\sum_{G\in\mathcal{G}}\mathcal{L}_1(G)
\end{equation}
\begin{equation}
    \mathcal{L}_2(G) = \frac{1}{\sum_{G\in \mathcal{G}}|{G}|}\sum_{G\in\mathcal{G}}\sum_{v\in G}\mathcal{L}_2(v)
\end{equation}
\begin{equation}
    \mathcal{L}_3(G) = \frac{1}{\sum_{G\in \mathcal{G}}|{G}|}\sum_{G\in\mathcal{G}}\sum_{v\in G}\mathcal{L}_3(v)
\end{equation}
\begin{equation}
    \mathcal{L}_4(G) = \frac{1}{|\mathcal{G}|}\sum_{G\in\mathcal{G}}\mathcal{L}_4(G)
\end{equation}
\begin{equation}
    \mathcal{L}_5(G) = \frac{1}{\sum_{G\in \mathcal{G}}|{G}|}\sum_{G\in\mathcal{G}}\sum_{v\in G}\mathcal{L}_5(v)
\end{equation}
where $|\mathcal{G}|$ denotes the number of graphs in batch $\mathcal{G}$, $|G|$ denotes the number of nodes in graph $G$.

we combine all the losses on graph-level and node-level to get a final loss function. Also, we introduce 4 hyper-parameters indicating the weight of each auxiliary task contributing to the major graph classification task. The final loss for a mini-batch of graphs $\mathcal{G}$ is defined as 
\begin{equation}
    \mathcal{L}(\mathcal{G}) = \mathcal{L}_1(\mathcal{G}) + \lambda_2 \mathcal{L}_2(\mathcal{G}) + \lambda_3\mathcal{L}_3(\mathcal{G}) + \lambda_4\mathcal{L}_4(\mathcal{G}) + \lambda_5\mathcal{L}_5(\mathcal{G})  
\end{equation}
where $\lambda_i$ denotes the weight of task $i$ for the major graph classification task.

\textbf{Discussion.}
Although our auxiliary tasks share some similar properties with the tasks in GNN node-level pre-training stage in \cite{hu2019strategies}, our multi-task graph learning framework is different from the self-supervised pretraining GNN framework in at least 3 aspects. (1) We have different workflows. Pretraining GNN follows the two-stage pretraining and fine-tuning framework, while our model does not have a pretraining process, containing only one training stage. (2) Pretraining GNN requires a large-scale graph dataset for pretraining, while our model does not. (3) For pretraining GNN, the node-level self-supervised pretraining and graph-level multi-task supervised pretraining are performed sequentially, and the fine-tuning process is performed only for the target task without any auxiliary tasks; while in our framework, the auxiliary tasks and the target task are trained jointly with a combined loss function. 

\section{Experiments}
In this section, we conduct experiments to evaluate the expressive power and generalizability of our model. The expressive power and generalizability correspond to the performance on training set and test set, respectively. 

\subsection{Experimental setup} \label{sec:expset}
\subsubsection{Datasets.} Four widely used benchmark datasets are involved in our experiments, including MUTAG, PTC, NCI1 and PROTEINS. A summary of these datasets is listed in Table \ref{tab:dataset}. We use one-hot encodings of node labels as the initial input features.

\begin{table*}[!t]
% \small
  \centering
    \begin{tabular}{l|c|c|c|c|c|c|l}
    \toprule
          & \#G & \#$C_G$ & \#$C_N$ & AvgN & AvgE  & MaxNeighb  & Source \\
\midrule       
  MUTAG &188 &	2 &	7 &17.93&	19.79 & 4  & \cite{debnath1991structure,kriege2012subgraph}\\
  PTC &344	&2 & 19	&14.29&	14.69 & 4 &  \cite{helma2001predictive,kriege2012subgraph}\\
  NCI1 &4110&	2	& 37  &29.87&	32.30 & 4  & \cite{wale2008comparison,shervashidze2011weisfeiler}\\
  PROTEINS &1113&	2 &  3	&39.06&	72.82 & 25 & \cite{borgwardt2005protein,dobson2003distinguishing}\\
    \bottomrule
    \end{tabular}%
  \caption{Dataset information. all the Dataset can be available at \url{https://ls11-www.cs.tu-dortmund.de/staff/morris/graphkerneldatasets}. \#G=number of graphs. \#$C_G$=number of graph classes. \#$C_N$=number of node classes. AvgN=average number of nodes in one graph. AvgE=average number of edges in one graph. MaxNeighb is the max 1-hop neighbors in all the nodes. }
  \label{tab:dataset}%
\end{table*}%

\subsubsection{Implementation details.} 
Since the initial node features are one-hot encodings, the summation with identical transformation is injective, thus we set whether the transformation function in the first layer is identical or an MLP as an optional hyperparameter. We set 5 GNN layers in our model, all MLPs contained in GNN layers or for prediction in the model had 1 hidden layer. In every hidden layer, we applied a batch normalization followed by a ReLU activation function. The Adam optimizer with learning rate decayed by 0.5 every 50 epochs. For each dataset, we trained the model for 300 epochs and recorded the performances on training set and validation set for each epoch. The batch size is 32. The drop out ratio is 0. The search space of hyper-parameters we tuned for each dataset are: (1) the number of hidden units \{16, 32, 64\}; (2) initial learning rate \{0.01, 0.001\}; {3} the transformation function in the first layer \{identical, MLP\}; (4) the weight of the 4 auxiliary tasks $\lambda_2,\lambda_3,\lambda_4,\lambda_5 \in \{0,1\}$. For each dataset, we followed the standard 10-fold cross validation protocol and splits from \cite{xu2019powerful}. Following the previous work \cite{xu2019powerful,maron2019provably,mao2020towards}, we reported the best averaged validation accuracy across the 10 folds for a fair comparison. Our experiments were run in 10 Tesla V100 GPUs.

\subsubsection{Compared Methods.}
We compared our model with a number of state-of-the-art methods listed in the first column in Table \ref{tab:result} for graph classification. Note that ExpGNN \cite{mao2020towards} and MTRL \cite{xie2020multi} fall in our framework. If the weights of all auxiliary tasks are 0, our model reduces to ExpGNN; if we only consider the node label preserving task, our model reduces to MTRL with ExpGNN as backbone. Besides, the compared methods also include the latest work on deep learning on graphs (e.g., \cite{maron2019provably,wang2020haar,xu2019powerful,morris2019weisfeiler}) and classical graph kernel methods (e.g., \cite{shervashidze2011weisfeiler,shervashidze2009efficient,yanardag2015deep}). Please refer to the corresponding papers for detailed introductions of the methods. For ExpGNN and MTRL we ran the experiments to get the results in our settings. For other baselines, we report the accuracy results in the original papers.

\subsection{Results}
We evaluate our model on both training set and test set. The performances on test set reflect the generalizability of a model, and the performance on training set usually reflects the expressive power of a model.

\subsubsection{Test performance.}
Table \ref{tab:result} lists the accuracies of our model compared with the baselines. The hyper-parameters that achieve the listed performances are attached in Table \ref{tab:hyperparameter} for reproduction. From Table \ref{tab:result}, while no one method can achieve the best for all the 4 datasets, no other models except our model DP-GNN can achieve the top 3 among the 21 methods for all the 4 datasets, and DP-GNN performs the best on PTC dataset. DP-GNN can consistently perform better than message passing GNN models that represent high expressive power, such as ExpGNN, MTRL, GIN, validating the efficacy of the distribution preserving task in our framework. 

\begin{table}[!t]

\scriptsize
  \centering
  \caption{Accuracy (\%) for graph classification with 10-fold cross validation (mean$\pm$std). The best performances are underlined. Top 3 performances on each dataset are bolded.}
    \begin{tabular}{l|c|c|c|c}
    \toprule
      Models    & MUTAG & PTC   & NCI1  & PROTEINS \\
    \midrule
    DP-GNN (ours) & \textbf{91.1$\pm$6.2} & \underline{\textbf{67.3$\pm$6.9}} & \textbf{84.2$\pm$1.9} & \textbf{76.9$\pm$5.3} \\
    ExpGNN \cite{mao2020towards} & 90.0$\pm$6.9 & 65.6$\pm$4.8 & 83.5$\pm$2.2 & 76.5$\pm$4.5 \\
    MTRL  \cite{xie2020multi}  & 90.0$\pm$6.0 & 65.0$\pm$7.4 & \textbf{83.7$\pm$2.3} & 76.5$\pm$5.0 \\
    GIN  \cite{xu2019powerful}  & 89.4$\pm$5.6 & 64.6$\pm$7.0 & 82.7$\pm$1.7 & 76.2$\pm$2.8 \\
    GCN  \cite{kipf2017semi}  & 85.6$\pm$5.8 & 64.2$\pm$4.3 & 80.2$\pm$2.0 & 76.0$\pm$3.2 \\
    GraphSAGE  \cite{hamilton2017inductive} & 85.1$\pm$7.6 & 63.9$\pm$7.7 & 77.7$\pm$1.5 & 75.9$\pm$3.2 \\
    3WL GNN  \cite{maron2019provably} & 90.5$\pm$8.7 & \textbf{66.2$\pm$6.5} & 83.2$\pm$1.1 & \textbf{77.2$\pm$4.7} \\
    HaarPool  \cite{wang2020haar} & 90.0$\pm$3.6 &   -    & 78.6$\pm$0.5 & \underline{\textbf{80.4$\pm$1.8}}\\
    PSCN  \cite{niepert2016learning} & \underline{\textbf{92.6$\pm$4.2}} & 60.0$\pm$4.8 & 78.6$\pm$1.9 & 75.9$\pm$2.8 \\
    DCNN  \cite{atwood2016diffusion}  & 67.0  & 56.6  & 62.6  & 61.3 \\
    DGCNN  \cite{zhang2018end} & 85.8$\pm$1.7 & 58.6$\pm$2.5 & 74.4$\pm$0.5 & 75.5$\pm$0.9 \\
    CapsGNN  \cite{xinyi2019capsule} & 86.7$\pm$6.9 & -     & 78.3$\pm$1.5 & 76.3$\pm$3.6 \\
    GCAPS-CNN  \cite{verma2018graph} & -     & \textbf{66.0$\pm$5.9} & 82.7$\pm$2.4 & 76.4$\pm$4.2 \\
    IEGN  \cite{Maron2019InvariantAE}  & 84.6$\pm$1 & 59.5$\pm$7.3 & 73.7$\pm$2.6 & 75.2$\pm$4.3 \\
    ECC  \cite{simonovsky2017dynamic}  & 76.11 &   -    & 76.82 & - \\
    1-2-3 GNN  \cite{morris2019weisfeiler} & 86.1  & 60.9  & 76.2  & 75.5 \\
    \midrule
    FGSD (Verma \cite{verma2017hunt}  & \textbf{92.12} & 62.8  & 79.8  & 73.42 \\
    WL  \cite{shervashidze2011weisfeiler}   & 83.8$\pm$1.5 &   -    & \underline{\textbf{84.5$\pm$0.4}} & - \\
    GK  \cite{shervashidze2009efficient}   & 81.6$\pm$2.1 & 57.3$\pm$1.4 & 62.5$\pm$0.3 & 71.7$\pm$0.5 \\
    Graph2vec  \cite{narayanan2017graph2vec} & 83.1$\pm$9.2 & 60.2$\pm$6.9 & 73.2$\pm$1.8 & 73.3$\pm$2.0 \\
    DGK \cite{yanardag2015deep}  & 87.4$\pm$2.7 & 60.1$\pm$2.5 & 80.3$\pm$0.5 & 75.7$\pm$0.5 \\
    \bottomrule
    \end{tabular}%

  \label{tab:result}%
\end{table}%

\begin{table}[htbp]
\small
  \centering
  \caption{The hyper-parameters corresponding to the best performance}
    \begin{tabular}{l|c|c|c|c}
    \toprule
          & MUTAG & PTC   & NCI1  & PROTEINS \\
          \midrule
    hidden dim & 32    & 64    & 32    & 16 \\
    learning rate & 0.01  & 0.01  & 0.001 & 0.01 \\
    first TF & identical & identical & MLP   & identical \\
    \midrule
    $\lambda_2$ & 1     & 0     & 1     & 1 \\
    $\lambda_3$ & 0     & 0     & 1     & 1 \\
    $\lambda_4$ & 0     & 1     & 1     & 0 \\
    $\lambda_5$ & 1     & 1     & 0     & 1 \\
    \midrule
    accuracy & 0.9111 & 0.6735 & 0.8421 & 0.7694 \\
    \bottomrule
    \end{tabular}%
  \label{tab:hyperparameter}%
\end{table}%

\subsubsection{Training curve.}
\begin{figure}[!t]
  \centering
  \subfloat[MUTAG]{\includegraphics[width=.45\linewidth,page=1]{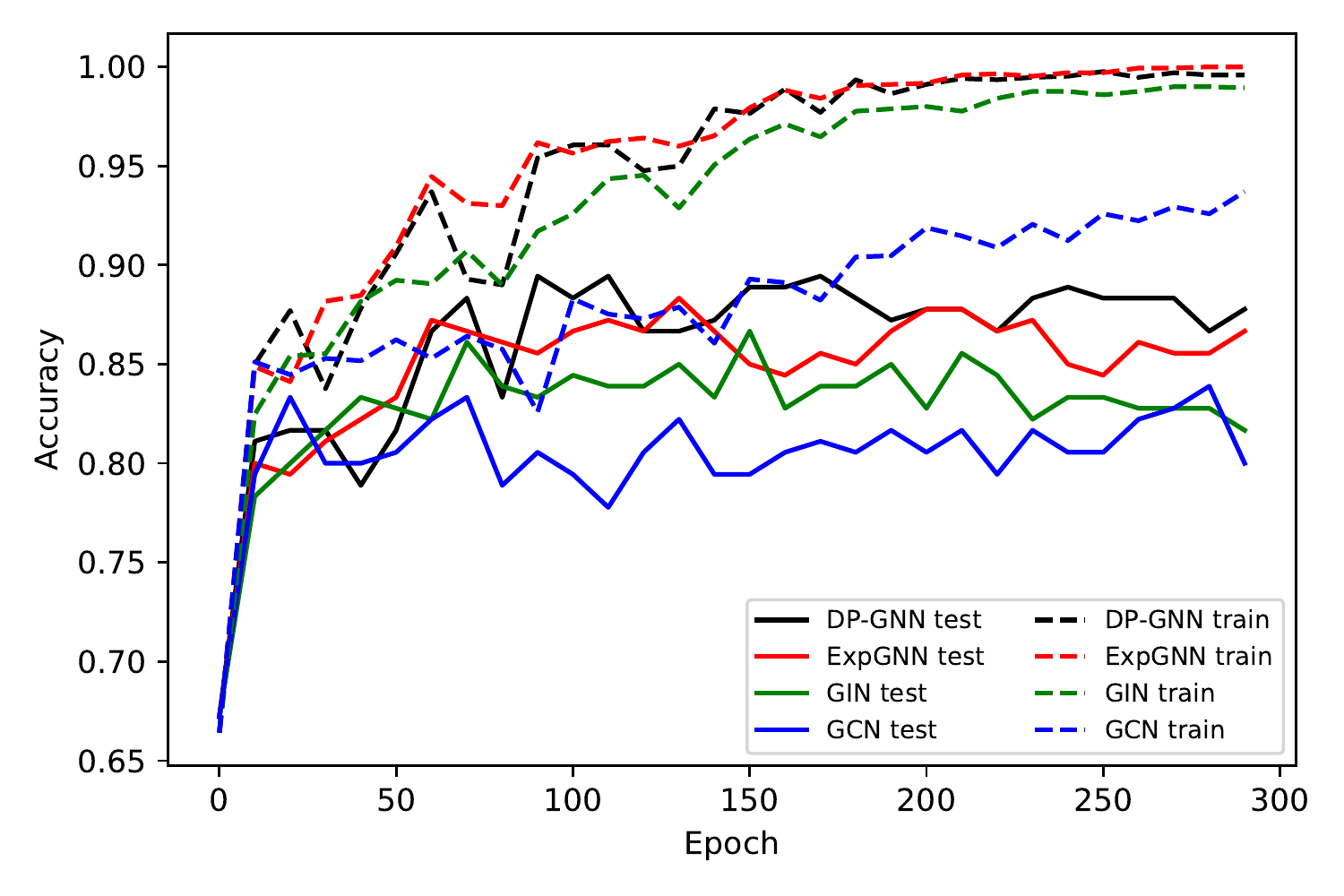}\label{fig:mutagcurve}}
  \subfloat[PTC]{\includegraphics[width=.45\linewidth,page=1]{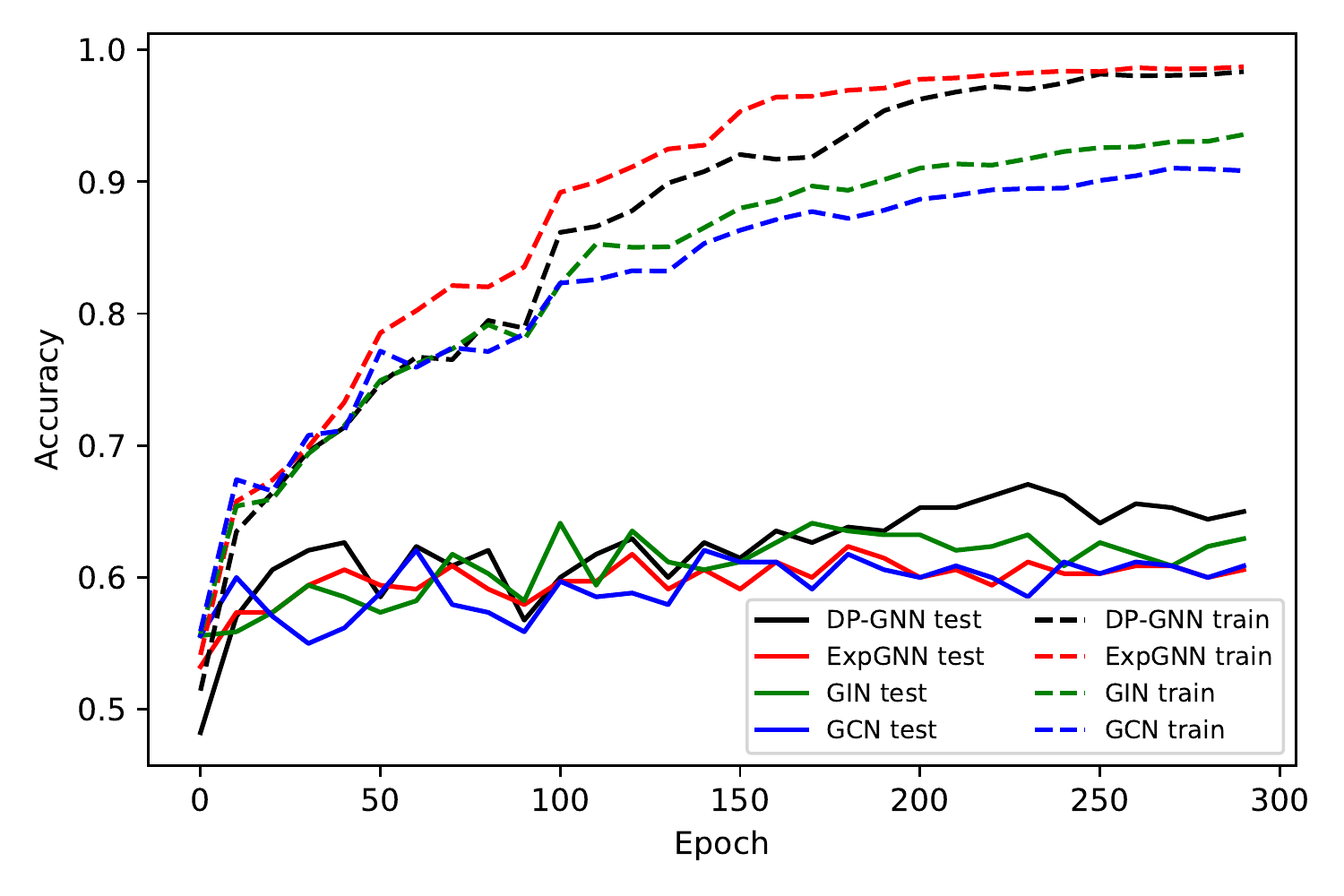}\label{fig:ptccurve}}
  \caption{The accuracy varying with the training epoch on training set and test set in the training process.}
  \label{fig:trainingcurve}
\end{figure}

We also averaged the accuracies on both training set and validation set across the 10 folds for each epoch to show how the performances changed in the training process. Figure \ref{fig:trainingcurve} illustrates how the accuracy vary along with the training epoch on both training set and test set. From Figure \ref{fig:trainingcurve}, on both MUTAG dataset (\ref{fig:mutagcurve}) and PTC dataset (\ref{fig:ptccurve}), the training accuracies of DP-GNN and ExpGNN are approaching 1 in the final epoch, validating the high expressive power of DP-GNN and ExpGNN. Also, Figure \ref{fig:trainingcurve} demonstrates the rank of models in expressive power is ExpGNN$\approx$DP-GNN$>$GIN$>$GCN, which is consistent with the theoretical analysis about expressive power \cite{mao2020towards,xu2019powerful}. DP-GNN is of high expressive power because DP-GNN use the backbone of ExpGNN. For the test accuracy, DP-GNN consistently performs best after a certain epoch when the model are trained adequately for both MUTAG and PTC datasets, validating the generalizability of DP-GNN.

\subsubsection{TSNE visualization.}
\begin{figure}[!t]
  \centering
  \subfloat[DP-GNN]{\includegraphics[width=.25\linewidth,page=1]{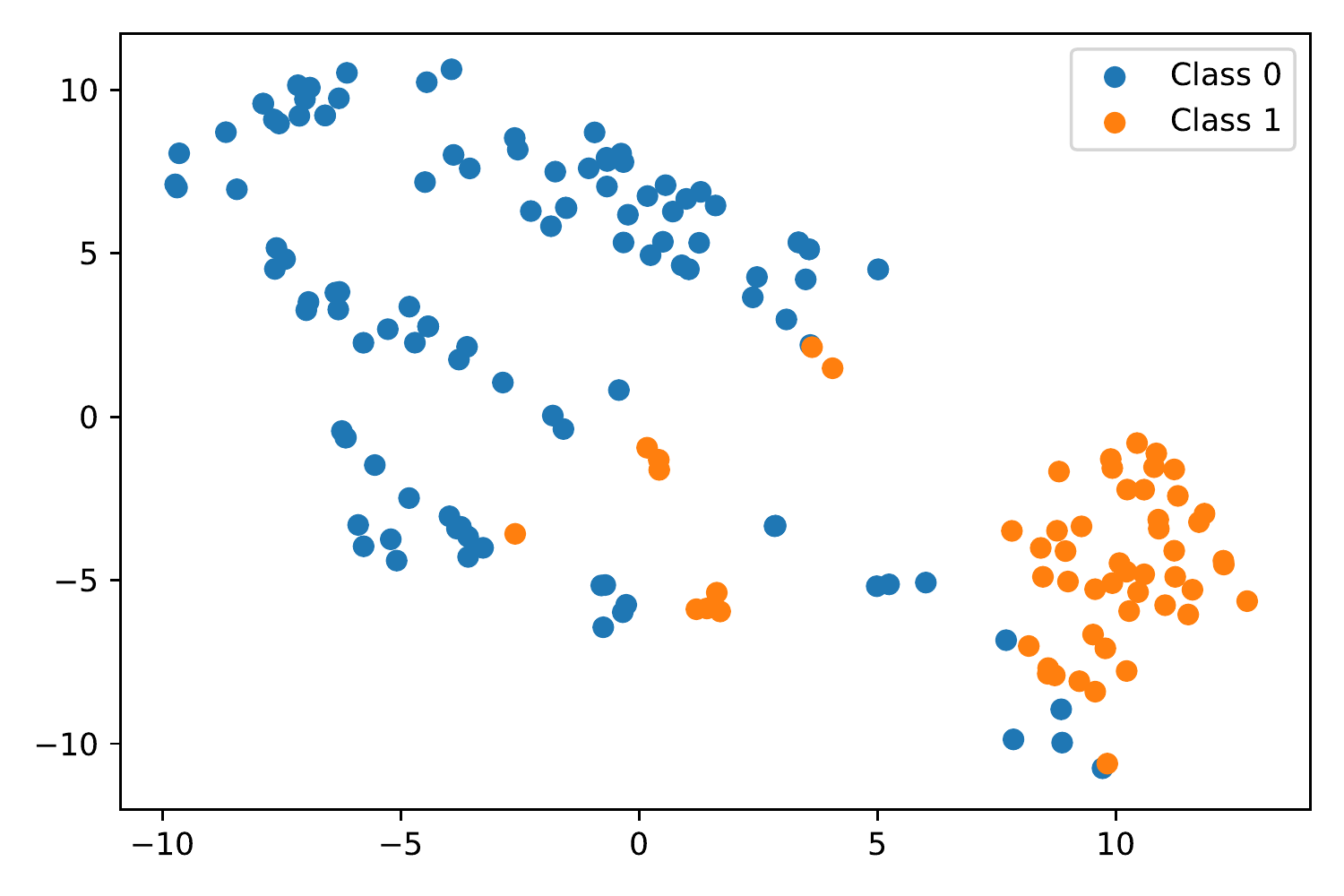}\label{fig:tsnedpgnn}}
  \subfloat[ExpGNN]{\includegraphics[width=.25\linewidth,page=1]{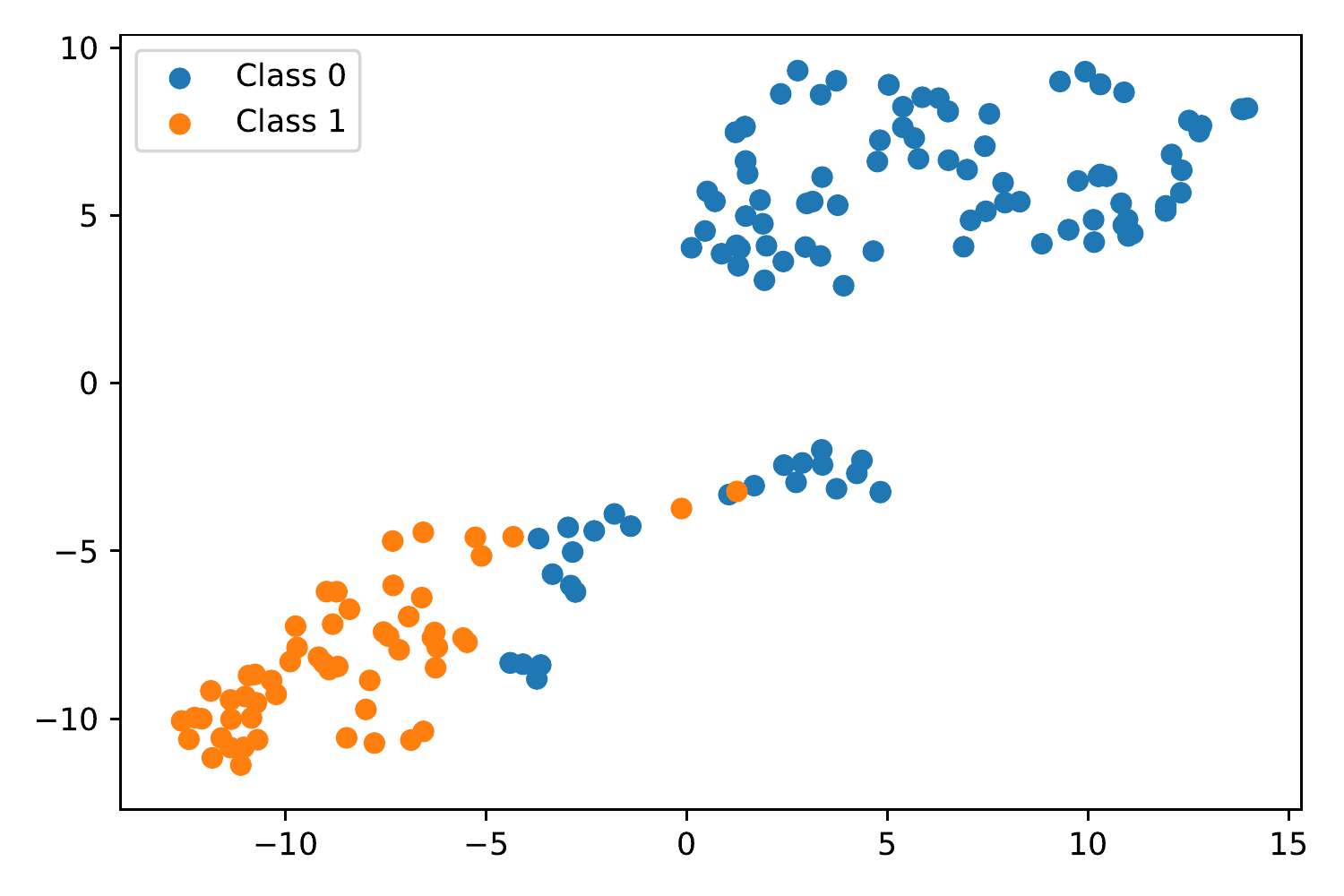}\label{fig:tsneexpgnn}}
%   \caption{t-SNE visualization of the output embeddings of MUTAG dataset.}
  \subfloat[DP-GNN]{\includegraphics[width=.25\linewidth,page=1]{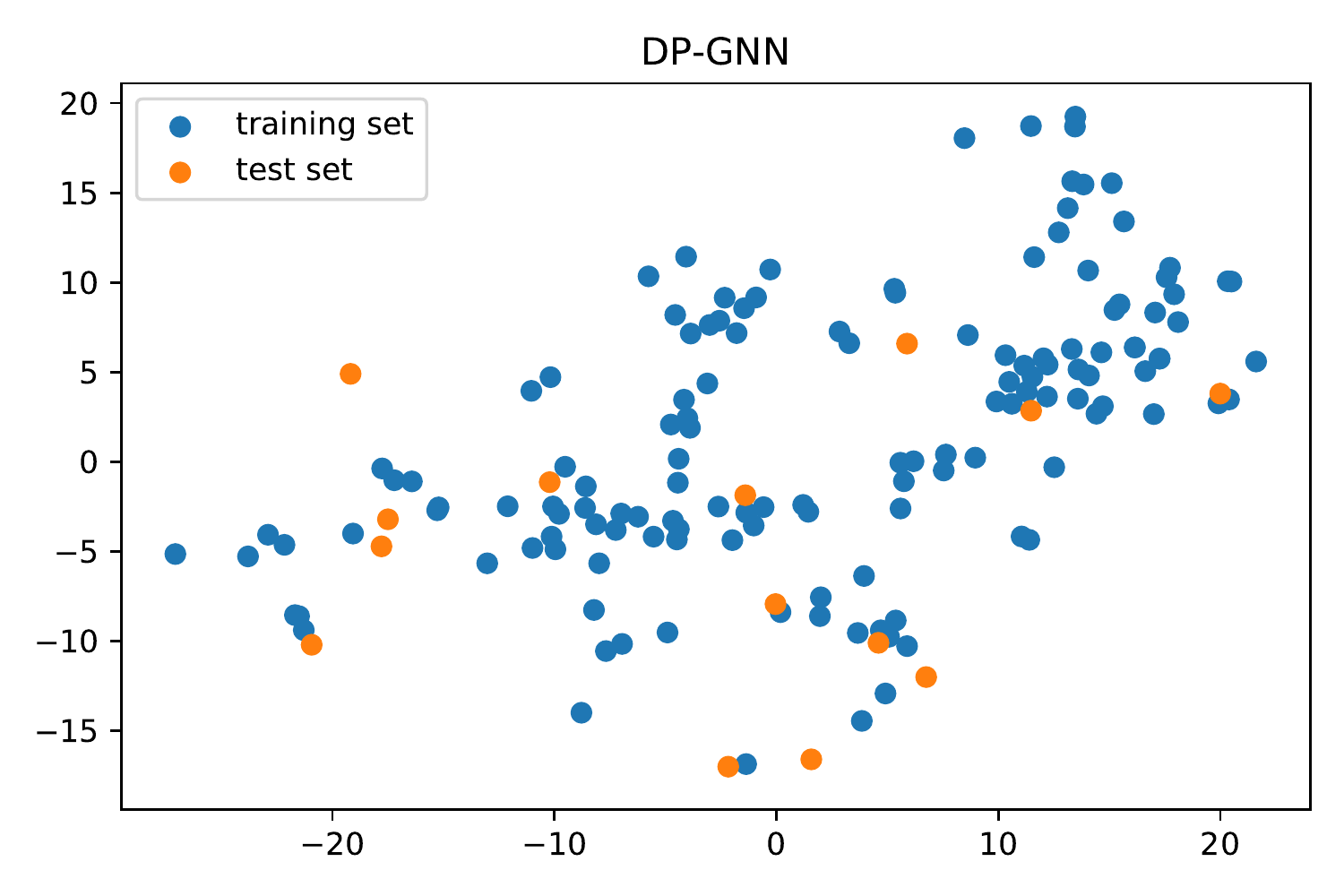}\label{fig:tsnedpgnn_trte}}
  \subfloat[ExpGNN]{\includegraphics[width=.25\linewidth,page=1]{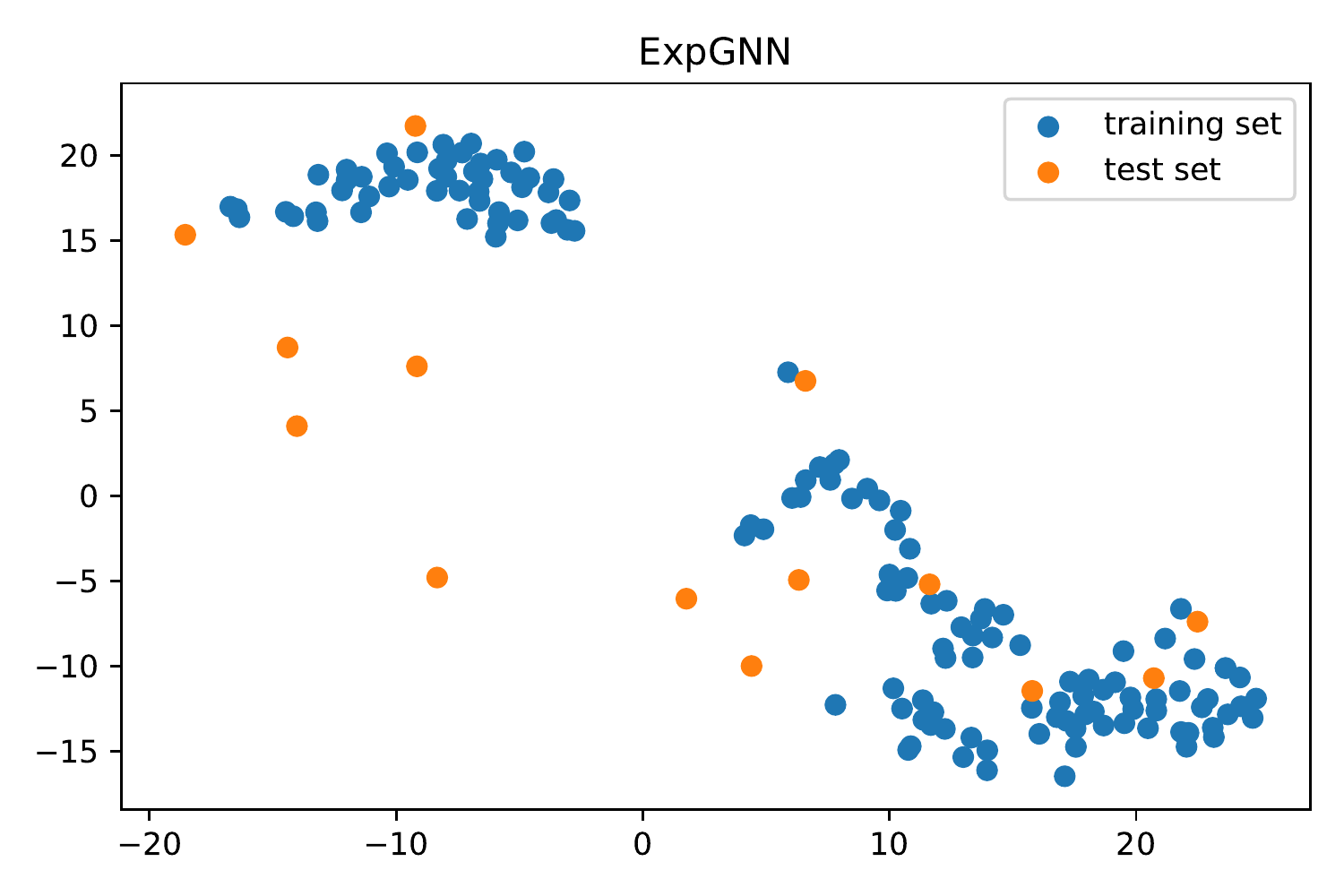}\label{fig:tsneexpgnn_trte}}
  \caption{t-SNE visualizations. (a-b) t-SNE visualization of different classes in MUTAG dataset; (c-d) t-SNE visualization of training set and test set in PTC dataset.}
  \label{fig:tsnevisual}
\end{figure}

We also showed the t-SNE visualization of the generated embeddings for DP-GNN and ExpGNN on MUTAG dataset in Figure \ref{fig:tsnevisual}. While both representations are visually separable for graph classification, we find that the representations generated by ExpGNN \ref{fig:tsneexpgnn} are more compact than by DP-GNN \ref{fig:tsnedpgnn}. Because the representations generated by ExpGNN were trained by optimizing the major graph classification tasks, and the representations of DP-GNN were trained by optimizing multiple auxiliary tasks as well as the major graph classification tasks, the representations of DP-GNN should contain more information that could used for the auxiliary tasks, thus not as compact as ExpGNN. However, compact representations on training set cannot ensure the compactness in test set, and usually cause more distribution shift. Non-compact and easily separable representations could have better generalizability.

To verify the distribution shift between training set and test set, we also plot the TSNE visualization of DP-GNN and ExpGNN on training set and test set for PTC dataset in Figure \ref{fig:tsnedpgnn_trte} and \ref{fig:tsneexpgnn_trte}, where DP-GNN indeed shows less distribution shift between training set and test set than ExpGNN.

\subsection{Ablation Study}
We recognize that for different graph classification tasks, not all the auxiliary tasks are helpful. For example, if the graphs are classified by the node numbers or if all the graphs contain only one kind of nodes, then task 2, i.e., node label preserving can do nothing with the major task. To determine what tasks are helpful to the major graph classification tasks on different datasets, we have done an ablation study. In this study, for each of the 4 auxiliary tasks, we run the model to determine if the single auxiliary task contributes to the major task independently, and then we combine all the helpful tasks for another run to see if the combination of helpful tasks contributes more to the performance of the major task. The results are listed in Table \ref{tab:ablation}, where we can see that the combination of independently helpful tasks do contribute more to the performance of the major task. Specifically, Task 2 and 5 are helpful on MUTAG and PROTEINS datasets; Task 4 and 5 are helpful on PTC dataset; Task 2, 3 and 4 are helpful on NCI1 dataset. We also list the best accuracy and the corresponding $\lambda$ in the last row of Table \ref{tab:ablation} for comparison.

\begin{table}[!t]
\small
  \centering
  \caption{The weights of auxiliary tasks relevant to the major task.$\lambda=[\lambda_2,\lambda_3,\lambda_4,\lambda_5]$}
    \begin{tabular}{c|c|c|c|c}
    \toprule
    $\lambda$ & MUTAG & PTC   & NCI1  & PROTEINS \\
    \midrule
    $[0,0,0,0]$ & 0.9000 & 0.6558 & 0.8348 & 0.7649 \\
    \midrule
    $[1,0,0,0]$ & \textbf{0.9000} & 0.6500 & \textbf{0.8375} & \textbf{0.7649} \\
    $[0,1,0,0]$ & 0.8889 & 0.6500 & \textbf{0.8350} & 0.7595 \\
    $[0,0,1,0]$ & 0.8944 & \textbf{0.6618} & \textbf{0.8367} & 0.7586 \\
    $[0,0,0,1]$ & \textbf{0.9056} & \textbf{0.6735} & 0.8343 & \textbf{0.7667} \\
    \midrule
    combined $\lambda$ & $[1,0,0,1]$ & $[0,0,1,1]$ & $[1,1,1,0]$ & $[1,0,0,1]$ \\
    accuracy & \textbf{0.9111} & \textbf{0.6735} & \textbf{0.8421} & \textbf{0.7685} \\
    \midrule
    best $\lambda$ & $[1,0,0,1]$ & $[0,0,1,1]$ & $[1,1,1,0]$ & $[1,1,0,1]$ \\
    best accuracy & \textbf{0.9111} & \textbf{0.6735} & \textbf{0.8421} & \textbf{0.7694} \\

    \bottomrule
    \end{tabular}%

  \label{tab:ablation}%
\end{table}%

\section{Conclusion}
In this paper, we propose DP-GNN to improve the generalizability of expressive GNN models by preserving several kinds of distribution information in the learned graph and node representations. The model is implemented by a multi-task learning framework where 4 auxiliary tasks are employed for distribution preservation. Since DP-GNN is built based on a highly-expressive GNN model, it can also achieve high expressive power. We validate the proposed DP-GNN for graph classification on multiple benchmark datasets. The accuracy performances of DP-GNN on training set are approaching 1, confirming the high expressive power of DP-GNN. The experimental results on test set demonstrate that our model achieves state-of-the-art performances on most of the benchmarks, validating the generalizability.

\bibliographystyle{named}
\bibliography{dpgnn}

\end{document}